\documentclass[letterpaper]{article} 
\usepackage{aaai2026}  
\usepackage{times}  
\usepackage{helvet}  
\usepackage{courier}  
\usepackage[hyphens]{url}  
\usepackage{graphicx} 
\usepackage{natbib}  
\usepackage{caption} 
\usepackage{algorithm}
\usepackage{algorithmic}

\usepackage{graphicx,verbatim}
\usepackage{booktabs}
\usepackage{multirow, float}
\usepackage{pifont, amssymb, amsmath}
\newcommand{\cmark}{\ding{51}}
\newcommand{\xmark}{\ding{55}}

\usepackage{mathtools}    
\usepackage{amsthm}       

\usepackage{amsfonts} 
\usepackage{enumitem}

\usepackage{newfloat}
\usepackage{listings}
\DeclareCaptionStyle{ruled}{labelfont=normalfont,labelsep=colon,strut=off} 
\floatstyle{ruled}
\newfloat{listing}{tb}{lst}{}
\floatname{listing}{Listing}
\title{DiA-gnostic VLVAE: Disentangled Alignment-Constrained Vision Language Variational AutoEncoder for Robust Radiology Reporting with Missing Modalities}
\author{
    Nagur Shareef Shaik\textsuperscript{\rm 1},
    Teja Krishna Cherukuri\textsuperscript{\rm 1},
    Adnan Masood\textsuperscript{\rm 2},
    Dong Hye Ye\textsuperscript{\rm 1},
}
\affiliations{
    \textsuperscript{\rm 1}Department of Computer Science, Georgia State University, Atlanta, GA, USA; 
    \textsuperscript{\rm 2}UST, Aliso Viejo, CA, USA.
    nshaik3@student.gsu.edu, tcherukuri1@student.gsu.edu, amasood@amp207.hbs.edu, dongye@gsu.edu
}

\usepackage{bibentry}

\begin{document}
\maketitle
\begin{abstract}
The integration of medical images with clinical context is essential for generating accurate and clinically interpretable radiology reports. However, current automated methods often rely on resource-heavy Large Language Models (LLMs) or static knowledge graphs and struggle with two fundamental challenges in real-world clinical data: (1) \textbf{missing modalities}, such as incomplete clinical context , and (2) \textbf{feature entanglement}, where mixed modality-specific and shared information leads to suboptimal fusion and clinically unfaithful hallucinated findings. To address these challenges, we propose the \textbf{DiA-gnostic VLVAE}, which achieves robust radiology reporting through \textbf{Di}sentangled \textbf{A}lignment. Our framework is designed to be resilient to missing modalities by disentangling shared and modality-specific features using a Mixture-of-Experts (MoE) based Vision-Language Variational Autoencoder (VLVAE). A constrained optimization objective enforces orthogonality and alignment between these latent representations to prevent suboptimal fusion. A compact LLaMA-\textit{X} decoder then uses these disentangled representations to generate reports efficiently. On the IU X-Ray and MIMIC-CXR datasets, DiA has achieved competetive BLEU@4 scores of 0.266 and 0.134, respectively. Experimental results show that the proposed method significantly outperforms state-of-the-art models.
\end{abstract}

\section{Introduction}
Radiology report generation (RRG) is a critical task in medical imaging that aims to produce accurate and comprehensive reports from scans, which can help lessen the burden on radiologists. Despite progress in computer vision and natural language processing, RRG remains a significant challenge due to the need for precise clinical insight and coherent report synthesis. This is often complicated by imbalanced datasets where rare conditions are underrepresented, which can compromise diagnostic reliability~\cite{yu2025adapter}.

Early models, such as R2Gen~\cite{chen2020generating} and CvT2Dis~\cite{nicolson2023improving}, relied exclusively on image features, using transformers and contrastive learning to refine visual representations. However, this image-centric approach has difficulty capturing nuanced diseases and integrating clinical reasoning. Subsequent efforts focused on improving vision-language integration. For example, XProNet utilized cross-modal prototypes for alignment~\cite{wang2022cross}, while METransformer used multiple learnable expert tokens to enhance textual consistency~\cite{wang2023metransformer}. Still, these models' reliance on image-centric patterns can lead to semantic discrepancies and clinical errors, especially when radiographic features of different diseases overlap, due to a lack of contextual grounding.

To address these limitations, recent models have begun to incorporate diagnostic context, such as disease pseudo-labels, knowledge graphs, or prior findings. Knowledge-driven approaches like MKSG~\cite{yang2022knowledge} and M2KT~\cite{yang2023radiology} use medical knowledge graphs to improve factual accuracy. Context-aware models such as KiUT~\cite{huang2023kiut}, DCL~\cite{li2023dynamic}, EKAGen~\cite{bu2024instance}, and PromptMRG~\cite{jin2024promptmrg} have also integrated expert knowledge and prior reports through graphs and prompts. While these methods enhance the clinical relevance of the generated reports, they have several technical constraints. For instance, they often lack explicit disentanglement, making it difficult to separate modality-specific knowledge from shared information. Consequently, the absence of context can lead to incomplete reports due to inefficient multi-modal alignment. Additionally, prompt-based models often depend on templates constructed from pseudo-diagnoses, which limits their adaptability and can significantly increase computational overhead due to their use of Large Language Models (LLMs).

Retrieval-augmented methods like SEI~\cite{liu2024structural} have advanced this area by extracting ``factual entities'' from a study, retrieving similar past cases, and using them to guide a cross-modal fusion decoder. However, this approach has its own issues. The entity-extraction and retrieval stages can be brittle, and the fusion network does not enforce explicit modality disentanglement or probabilistic feature gating. This leaves the model vulnerable to feature interference within what the authors term an ``unstable fusion space''. Furthermore, when contextual information is missing, these models often fall back on deterministic rules instead of a principled probabilistic strategy, which can cause errors from earlier stages to propagate.

To tackle these challenges, we introduce the \textbf{DiA-gnostic VLVAE}, designed for \textbf{robust radiology reporting} by leveraging the principle of \textbf{Disentangled Alignment}. To handle missing modalities and dynamic patient states, the framework uses real-time clinical data, including demographics, symptoms, and prior history, as dynamic context. Its core is a Vision-Language Variational Autoencoder \cite{mao2023multimodal} that disentangles modality-specific and shared latent representations, ensuring consistent vision-language alignment even when context is incomplete. This is supported by a Vision-Language Representation Learning module using Guided Context Attention \cite{Cherukuri2024Guided} and a Modality Abstractor \cite{vaswani2017attention} for effective cross-modal feature fusion. Finally, a compact and efficient LLaMA-\textit{X} decoder generates clinically precise reports, avoiding the template rigidity of prompt-based models \cite{jin2024promptmrg} while outperforming more resource-intensive alternatives in adaptability and computational efficiency.

\section{Related Work}
\label{sec:related_work}

\subsubsection{Fusion of Heterogeneous Medical Data} 
Fusing heterogeneous medical data, such as EHR, clinical notes, and various medical imaging types \cite{venugopalan2021multimodal, mohsen2022artificial}, has shown significant potential for improving clinical tasks like prognosis prediction \cite{kline2022multimodal, cheerla2019deep}, phenotyping \cite{hayat2022medfuse}, and medical image segmentation \cite{huang2020multimodal}. This integration of diverse data sources is a clear trend aimed at building more comprehensive and accurate clinical models \cite{huang2020fusion}.

\subsubsection{Handling Missing Modality} 
In practice, some clinical data modalities are inevitably missing \cite{huang2020fusion}. A common solution is late fusion, where predictions from independently modeled modalities are aggregated at the decision level \cite{yoo2019deep, steyaert2023multimodal}. However, this approach can be suboptimal as it fails to capture the interactions between modalities \cite{huang2020fusion}. More recent research has explored generative methods to impute or reconstruct missing data at the feature or instance level \cite{ma2021smil, zhang2022m3care, sharma2019missing}. These techniques may use a Bayesian meta-learning framework \cite{ma2021smil} or impute features in the latent space with auxiliary information \cite{zhang2022m3care}. Despite these advances, results from generated data may not be robust \cite{li2023multi-modality, drfuse2024}, and handling missing data in highly heterogeneous settings like image-and-text fusion remains an open challenge \cite{drfuse2024}.

\subsubsection{Disentangled Representation Learning} 
A promising approach for handling both missing data and modal inconsistency is to disentangle shared and modality-specific information \cite{drfuse2024, imdr2025, drim2024}. The goal is to learn representations that separate common, patient-related information from unique, modality-specific details \cite{drim2024}. This is often achieved by imposing explicit constraints on the latent space. Common techniques include enforcing orthogonality between shared and specific representations to minimize redundancy \cite{braman2021deep, drfuse2024} or minimizing their mutual information, often via an adversarial objective \cite{sanchez2020learning, imdr2025, drim2024}. Concurrently, the alignment of shared representations is enforced using methods like Jensen-Shannon divergence (JSD) \cite{drfuse2024} or contrastive objectives \cite{drim2024}. While most prior work focused on more homogeneous modalities like different MRI scans \cite{chen2019robust, shen2019brain}, 
DiA introduces a probabilistic tri‑factor decomposition that leverages a Vision–Language VAE with a shared‑gate Mixture‑of‑Experts and a unified Disentangled‑Alignment constraint, enabling robust radiology reporting from highly heterogeneous inputs with missing modalities.

\begin{figure*}[!t]
    \centering
    \vspace{-10pt}
    \includegraphics[width=.95\textwidth]{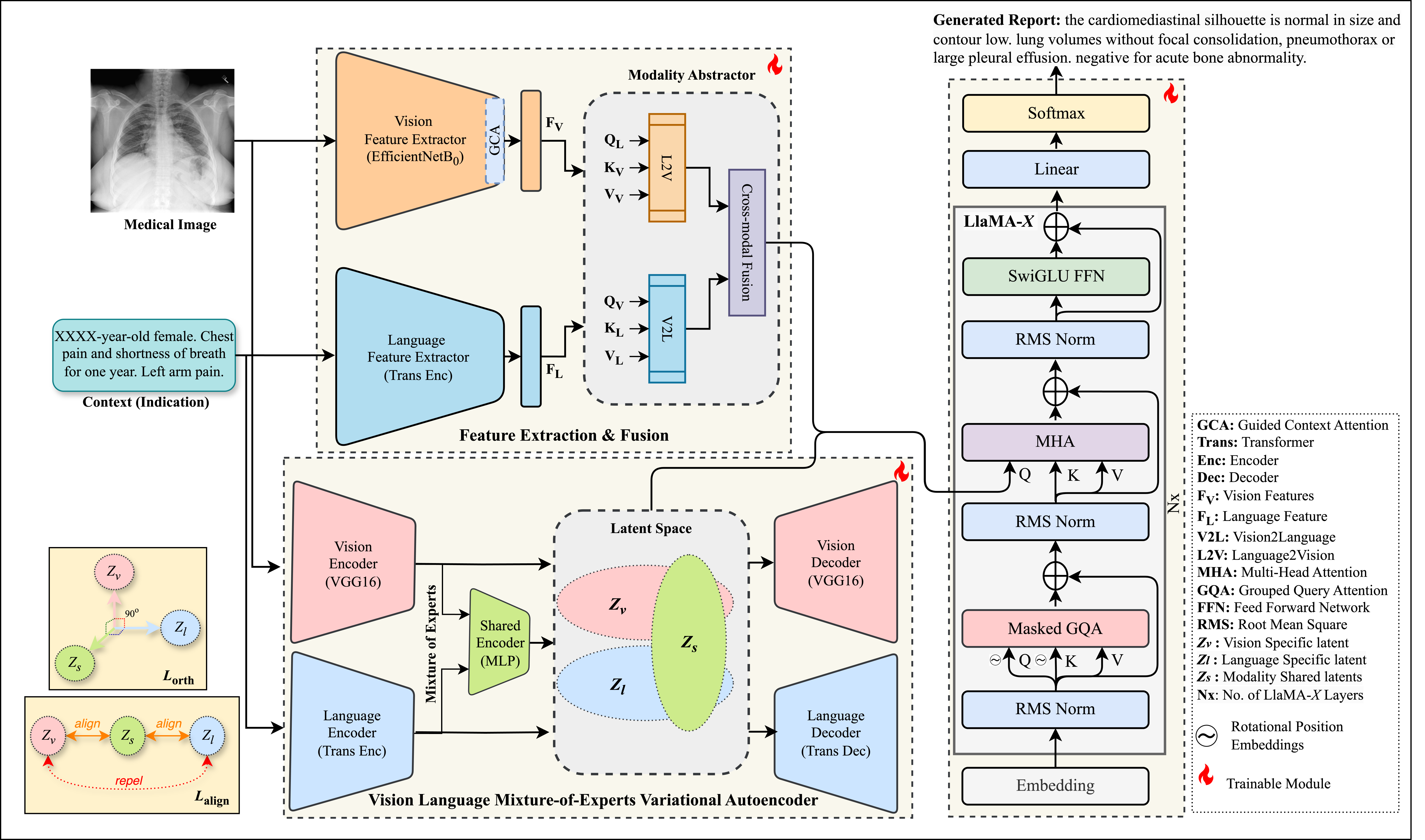}
    \vspace{-5pt}
    \caption{\textbf{Architecture of DiA:} Extracts vision features using \textit{EfficientNetB$_0$} with \textit{Guided Context Attention} and language features via a \textit{Transformer Encoder}, fused by a \textit{Modality Abstractor}; learns modality-specific latents ($ Z_v, Z_l $) using \textit{VAEs (VGG16 and Transformer)} and shared latent ($ Z_s $) through a \textit{Mixture-of-Experts Shared Encoder}, disentangled via $ \mathcal{L_\text{orth}} $, aligned with $ \mathcal{L_\text{align}} $; generate reports using LlaMA-\textit{X} Decoder.}
    \vspace{-10pt}
    \label{fig:diva-architecture}
\end{figure*}

\section{Methodology}
\label{sec:methods}
The \textbf{DiA-gnostic VLVAE} is a principled probabilistic approach for \textbf{robust radiology reporting} designed to be resilient to \textbf{missing modalities} such as incomplete clinical context. The framework is built on the principle of \textbf{Disentangled Alignment}, which it achieves by learning a tri-factor latent space that explicitly separates modality-specific (vision, language) features from shared cross-modal semantics. To handle missing data, the shared latent is inferred via a Mixture-of-Experts (MoE) posterior, a theoretically grounded method that allows the model to marginalize an absent expert while preserving inferential integrity. This factorization is guided by a dual-consistency constraint: an \textbf{orthogonality} term disentangles the latent factors, while a \textbf{contrastive alignment} term ensures the shared space is predictive of each modality, leading to robust and faithful generation. This disentangled structure is learned by our novel Vision-Language Mixture-of-Experts Variational Auto-Encoder (VL-MoE-VAE) module and is used to drive report generation through an efficient LLaMA-\textit{X} decoder.

\subsection{Problem Formulation}
\label{subsec:problem}
Let our dataset be $\mathcal{D} = \{ (V_i, L_i, R_i) \}_{i=1}^N$, where for each subject $i$, $V_i \in \mathbb{R}^{H \times W \times C}$ represents a medical image (e.g., Chest X-Ray), $L_i = \{ l_{i,k} \}_{i=1}^{_{K_i}}$ captures clinical indications (e.g., patient demographics, symptoms, prior history) with $K$ elements, and $R_i = \{ r_{i,t} \}_{t=1}^{_{T_i}}$ is the corresponding radiology report. Our primary objective is to learn a conditional generative model $p(R \mid V, L)$ that maximizes the likelihood of producing the correct report $R$ given the image $V$ and the accompanying clinical context $L$. A critical principle for achieving robust reporting is \emph{modality resilience}: the framework must remain effective even when one modality is absent, particularly the clinical context $L$. Consequently, the framework must also support principled inference for the marginal scenario
$p(R \mid V)$.

\subsection{Feature Extraction and Fusion}
\label{subsec:feature_extraction_fusion}
Before probabilistic modeling, we transform the raw, high-dimensional inputs into a unified, semantically rich feature space. This stage serves as a powerful feature extraction baseline, complementing DiA.

\subsubsection{Vision \& Language Feature Extractor}
We leverage a pre-trained convolutional neural network, EfficientNetB$_0$ \cite{tan2019efficientnet}, to extract high-level features from input image $V$. To capture clinically relevant global patterns that are often missed by local receptive fields, we augment the backbone with a Guided Context Attention (GCA)~\cite{Cherukuri2024Guided} mechanism. This module produces a spatially-aware feature map that is projected into the final vision feature, $ F_V \in \mathbb{R}^{S_V \times E} $, where $ S_V $ captures spatial dimensions and $ E $ is the number of feature channels. The clinical context $L$ is tokenized and processed by a standard Transformer encoder~\cite{vaswani2017attention} to capture complex semantic relationships, producing a sequence of contextualized embeddings $ F_{L} \in \mathbb{R}^{S_L \times E} $, where $ S_L $ is the maximum sequence length.

\subsubsection{Modality Abstractor}
To align and integrate these heterogeneous features, we use a Modality Abstractor based on bidirectional cross-attention~\cite{vaswani2017attention}. First, the vision features $F_V$ and language features $F_L$ are projected into query (Q), key (K), and value (V) representations using learnable weight matrices. The module then allows features from each modality to query the other, dynamically highlighting visually-grounded clinical terms and text-relevant image regions. This process computes both vision-to-language $F_{V2L}$ and language-to-vision $F_{L2V}$ representations via multi-head attention:
\begin{align}
    F_{V2L} &= F_{V} + \text{Softmax}\left(\frac{Q_{V} \cdot K_L^\top}{\sqrt{d_{k}}}\right) \cdot V_L \\
    F_{L2V} &= F_L + \text{Softmax}\left(\frac{Q_L \cdot K_{V}^\top}{\sqrt{d_{k}}}\right) \cdot V_{V}
\end{align}
where $ d_{k} $ is the key vector’s dimension. The resulting features are concatenated to form a unified multi-modal representation $F_{VL}$, integrating complementary features for downstream VLVAE module. 


\subsection{Vision-Language Mixture-of-Experts VAE}
\label{subsec:va-moe-vae}
We formulate DiA's probabilistic framework using a Multi-modal Variational Autoencoder (MVAE) (see Fig. \ref{fig:diva-architecture}) that learns a \textbf{Tri-factor Latent Decomposition}. This decomposition is designed to disentangle the sources of variation in vision-language data into three distinct latent variables: a vision-specific latent $Z_v$, a language-specific latent $Z_l$, and a shared, cross-modal latent $Z_s$. As the true posterior over the latents, $p_\theta(Z_v, Z_l, Z_s | V, L)$, is intractable, we introduce a variational approximation with a specific factorization: $q_\phi(Z_v, Z_l, Z_s | V, L) \sim q_{\phi_v}(Z_v | V) \cdot q_{\phi_l}(Z_l | L) \cdot q_{\phi_s}(Z_s | V, L)$. Here, $q_{\phi_v}$ and $q_{\phi_l}$ are encoders for the modality-specific latents, while $q_{\phi_s}$ is a joint encoder for the shared latent, which uses a Mixture-of-Experts (MoE) strategy to ensure robustness against missing modalities. 

\subsubsection{Modality-Specific Latent Inference}
\label{subsubsec:factorization}

The model's structure is guided by its generative process, which assumes that each observed modality is generated independently from its corresponding specific latent variable. For the vision modality, a latent variable $Z_v$ is sampled from a prior distribution $p(Z_v)$, and the image is generated by a decoder $p_{\theta_v}(V \mid Z_v)$, parameterized by $\theta_v$. Similarly, the language latent $Z_l$ is sampled from its prior $p(Z_l)$ to generate the clinical context via $p_{\theta_l}(L \mid Z_l)$, with parameters $\theta_l$. This design introduces a critical inductive bias: all information necessary to reconstruct a modality must be encoded in its specific latent variable, which enforces representational independence and facilitates disentangled learning. 

To learn the parameters, we need to infer the values of the latent variables from the data. This requires computing the true posterior distributions, $p_{\theta_v}(Z_v\mid V)$ and $p_{\theta_l}(Z_l\mid L)$, which are intractable to compute directly. To overcome this, we employ variational inference, introducing encoder networks to approximate these true but intractable posteriors. The vision encoder, $q_{\phi_v}(Z_v \mid V)$, uses a pre-trained VGG16 network~\cite{simonyan2014very} followed by a fully connected layer to produce the Gaussian parameters $(\mu_v, \sigma_v^2)$ for the approximate posterior over $Z_v$. The language encoder, $q_{\phi_l}(Z_l \mid L)$, is a Transformer-based encoder~\cite{liu2019transformer} that outputs $(\mu_l, \sigma_l^2)$ for the approximate posterior over the language-specific latent $Z_l$.

\subsubsection{Shared Latent Inference via Mixture-of-Experts}
\label{subsubsec:moe}
To model the shared latent variable $Z_s$, DiA employs a Mixture-of-Experts (MoE) strategy~\cite{shi2019variational} via a dedicated shared encoder. This approach contrasts with Product-of-Experts (PoE) approaches~\cite{wu2018multimodal}, which can produce overconfident posterior estimates and degrade significantly when a modality is missing. The MoE formulation provides a more robust alternative for learning from partially observed data.

The shared encoder approximates the posterior over $Z_s$ as a weighted combination of unimodal \textit{expert} posteriors. For each modality $M \in \{V, L\}$, the encoder outputs parameters $(\mu_{s}, \sigma_s^{2})$ and corresponding mixture weights $\pi_M$. The overall MoE posterior is then defined as:
\begin{equation}
q_{\phi_s}(Z_s \mid V, L) = \sum_{M \in \{V, L\}} \pi_M \cdot q_{\phi_s}(Z_s \mid M),
\label{eq:moe}
\end{equation}
where the mixture coefficients $\pi_M$ are non-negative and sum to one. This allows the model to adaptively the contribution of each modality to the shared representation. 

\subsubsection{Learning Objective}
\label{subsubsec:objective}
The overall learning objective for the proposed VL-MoE-VAE is to maximize the Evidence Lower Bound (ELBO)~\cite{mao2023multimodal} on the marginal log-likelihood. The ELBO balances accurate reconstruction with structured regularization over the latent space to enforce the desired disentangled alignment across $Z_v$, $Z_l$, and $Z_s$. The full objective is defined as:
\begin{align}
    &\mathcal{L}_{\text{ELBO}} = \mathbb{E}_{q_{\phi_s}(Z_s | V, L)} \Big[ \mathbb{E}_{q_{\phi_v}(Z_v | V)} \left[ \log p_{\theta_v}(V | Z_v) \right] \notag \\
    &\quad + \mathbb{E}_{q_{\phi_l}(Z_l | L)} \left[ \log p_{\theta_l}(L | Z_l) \right] \Big] \notag \\
    &\quad -\Big[ D_{\text{KL}}(q_{\phi_v}(Z_v | V) \| p(Z_v)) + D_{\text{KL}}(q_{\phi_l}(Z_l | L) \| p(Z_l)) \Big] \notag \\
    &\quad - \text{JSD}(q_{\phi_s}(Z_s | V, L) \, , \, p(Z_s))
    \label{eq:elbo}
\end{align}
This objective function evaluates the model's ability to reconstruct the input modalities $(V, L)$ from their respective specific latents $Z_v$ and $Z_l$, conditioned on a shared latent variable $Z_s$. It also encourages the modality-specific posteriors $q_{\phi_v}(Z_v \mid V)$ and $q_{\phi_l}(Z_l \mid L)$ to remain close to standard Gaussian priors $\mathcal{N}(0, I)$ via a Kullback-Leibler (KL) divergence penalty.

A key aspect of our Mixture-of-Experts (MoE) formulation is the use of Jensen-Shannon Divergence (JSD)~\cite{menendez1997jensen} to regularize the shared latent $Z_s$. Unlike the standard KL divergence, which can lead to \emph{component collapse} where only one expert contributes to the posterior~\cite{minka2005divergence}, the symmetric and bounded nature of JSD is more suitable for mixture distributions. It encourages the entire mixture to align with the prior, promoting stability and ensuring all experts contribute meaningfully to the shared latent representation, a choice consistent with recent findings in multimodal generative modeling~\cite{sutter2020multimodal, yao2024drfuse}.    

\subsection{Disentangled Alignment Constraint}
\label{subsec:da}
The ELBO objective alone does not guarantee that the latent factors are either semantically meaningful or disentangled. To explicitly enforce the desired properties of disentanglement between shared and modality-specific factors, and strong alignment within the shared space, we introduce a novel Disentangled Alignment Constraint, which combines two regularization terms detailed below.

\subsubsection{Orthogonality for Disentanglement}
\label{subsubsec:orth}
To promote statistical independence between modality-specific and shared latent representations, we introduce an orthogonality constraint on the latent space, a technique demonstrated to be effective in structured representation learning~\cite{bousmalis2016domain}. Specifically, we enforce uncorrelatedness between the latent variables $Z_v$, $Z_l$, and $Z_s$ by first applying a whitening transformation to each, resulting in zero-mean, unit-covariance representations denoted as $\bigl(\tilde{{Z}}_v,, \tilde{{Z}}_l,, \tilde{{Z}}_s\bigr)$. This is implemented via a batch normalization layer applied to each latent subspace. The orthogonality loss is then formulated as the sum of squared Frobenius norms of the pairwise cross-covariance matrices:
\begin{equation}
\mathcal{L}_{\text{orth}} = \|\tilde{Z}_s^{\top} \tilde{Z}_v\|_F^2 + \|\tilde{Z}_s^{\top} \tilde{Z}_l\|_F^2 + \|\tilde{Z}_v^{\top} \tilde{Z}_l\|_F^2
\end{equation}
Minimizing $\mathcal{L}_{\text{orth}}$ penalizes any statistical correlation between the latent subspaces, thereby encouraging disentanglement. This uncorrelation, when combined with whitening, approximates statistical independence under the assumption of non-Gaussianity, a core principle underlying Independent Component Analysis (ICA)~\cite{hyvarinen2001independent}.

\subsubsection{Contrastive Alignment of the Shared Space}
\label{subsubsec:align}
While orthogonality promotes statistical independence, it does not inherently guarantee the semantic relevance of the shared representation $Z_s$. To address this, we introduce a contrastive alignment objective based on the InfoNCE loss~\cite{rusak2024infonce}, which aligns $Z_s$ with the modality-specific latents $Z_v$ and $Z_l$. This objective encourages $Z_s$ to exhibit higher similarity with its corresponding modality-specific latent while treating the other as a negative sample. Formally, the alignment loss is defined as:
\begin{align}
    &\mathcal{L}_{\text{align}} = - \mathbb{E}_{q(Z_v, Z_s)} \left[ \log \frac{\exp(\text{sim}(Z_s, Z_v)/\tau)}{\sum_{Z'\in{\{Z_v,Z_l\}}} \exp(\text{sim}(Z_s, Z')/\tau)} \right] \nonumber \\
    & - \mathbb{E}_{q(Z_l, Z_s)} \left[ \log \frac{\exp(\text{sim}(Z_s, Z_l)/\tau)}{\sum_{Z'\in{\{Z_v,Z_l\}}} \exp(\text{sim}(Z_s, Z')/\tau)} \right]
    \label{eq:infonce}
\end{align}
where $\text{sim}(\cdot)$ denotes cosine similarity, and $\tau$ is a temperature parameter. This formulation ensures that $Z_s$ remains semantically coherent with both modalities. From an information-theoretic perspective, minimizing $\mathcal{L}_{\text{align}}$ effectively maximizes the mutual information between the shared and specific latents ($I(Z_s; Z_v)$ and $I(Z_s; Z_l)$), ensuring that the shared latent $Z_s$ captures semantic information common to both modalities~\cite{poole2019variational}. 

When combined, the orthogonality and alignment objective enable the model to learn latent spaces that are both statistically disentangled and semantically rich. This dual constraint is crucial for improving the model's generalization, robustness, and interpretability in multi-modal settings.

\subsection{LlaMA-\textit{X} Decoder}
The final report is generated by the LLaMA-\textit{X} Decoder, which is trained to model the dependencies between the report text and the fused multi-modal representations from the preceding modules. The entire DiA freamework is optimized end-to-end with a composite loss function. 

The LLaMA-\textit{X} Decoder is a compact adaptation of LLaMA~\cite{touvron2023llama}. It uses a GPT-derived Cross-Attention~\cite{brown2020language} to condition the report generation on the fused multi-modal representations from both the Modality Abstractor ($F_{VL}$) and VL-MoE-VAE ($Z_v, Z_l, Z_s$). The architecture incorporates several optimizations for efficiency and performance: (1) Rotary Positional Encodings (RoPE) which embed relative positional information via rotation matrices in the query and key vectors to efficiently handle long sequence lengths; (2) Grouped Query Attention which partitions queries into groups and leverages Key-Value (KV) caching to minimize redundant computations during inference; (3) SwiGLU Feed-Forward Network (FFN) that is defined as $ \text{SwiGLU}(x) = (x W_1) \odot \sigma(x W_2) W_3 $, with SiLU activation $ \sigma(\cdot)$ to enhance feature transformation and mitigate the vanishing gradient problem; (4) RMS Pre-Normalization that is defined as $ x' = x / \sqrt{\text{mean}(x^2) + \epsilon} $ to stabilize the inputs to the attention and feed-forward layers. 

The decoder is trained by optimizing a standard cross-entropy loss, $ \mathcal{L}_{\text{CE}} = - \sum_{i=1}^{N} \sum_{j=1}^{T} r_{ij} \log(\hat{r}_{ij}) $ to align predicted reports $\hat{r}$ with ground-truth $r$ over $ T $ tokens. The overall objective for the DiA framework integrates this generation loss with previously defined objectives for the VL-MoE-VAE and the Disentangled Alignment Constraint. The total loss is a weighted sum:
\begin{equation}
\mathcal{L}_{\text{total}} = \mathcal{L}_{\text{CE}} + \mathcal{L}_{\text{ELBO}} + \lambda_1 \mathcal{L}_{\text{orth}} + \lambda_2 \mathcal{L}_{\text{align}},
\end{equation}
where $\lambda_1$ and  $\lambda_2$ are hyperparameters that balance the contributions of the orthogonality and alignment losses, respectively. This composite objective ensures that the model learns to generate accurate reports while maintaining a robust, disentangles latent structure.

\subsection{Inference with Missing Context}
\label{sec:inference_missing}
A key advantage of the DiA framework is its inherent robustness to missing modalities, a common scenario in clinical workflows where the image \(V\) is present but the clinical context \(L\) may be absent. This resilience is a direct consequence of using a Mixture‑of‑Experts (MoE) posterior to infer the shared latent \(Z_s\). At inference time, if a modality \(L\) is unavailable,
a designated ``null'' token is passed to corresponding expert. As the MoE router was exposed to the same token during training, it learns to down‑weight the unavailable modality automatically, i.e.\ $\pi_L \!\approx\! 0$ and $\pi_V \!\approx\! 1$ in Eq.~\eqref{eq:moe}. This allows the posterior to gracefully reduce to being conditioned only on the available data $q_{\phi_s}(Z_s \mid V)$ without requiring any imputation or architectural changes. 


This process is theoretically sound. By substituting the reduced posterior into the training objective in eq. \eqref{eq:elbo} and discarding terms involving the missing modality \(L\), the objective becomes a marginal ELBO. 
\begin{align}
    &\mathcal{L}_{\text{ELBO}}^{(V)} = \mathbb{E}_{q_{\phi_v}(Z_v \mid V)} \bigl[\log p_{\theta_v}(V \mid Z_v)\bigr] \\
    &\quad - \,D_{\text{KL}}\!\bigl(q_{\phi_v}(Z_v \mid V)\,\|\,p(Z_v)\bigr) - \,\text{JSD} \bigl(q_{\phi_s}(Z_s \mid V),\,p(Z_s)\bigr)  \notag 
\end{align}

This new objective $\mathcal{L}_{\text{ELBO}}^{(V)}$ remains a valid lower bound on the marginal log-likelihood of the observed data ( $\mathcal{L}_{\text{ELBO}}^{(V)} \le \log p_\theta(V)$), ensuring the learning procedure is principled for any subset of modalities. 

The model's effective performance in this scenario stems from the contrastive alignment term \(\mathcal{L}_{\text{align}}\) applied during training. By maximizing the mutual information between the shared latent and each specific modality \(I(Z_s;Z_v)\) and \(I(Z_s;Z_l)\), the shared latent $Z_s$ learns to encode salient cross‑modal semantics. Consequently, even when inferred from a single modality, $Z_s$ still provides the LLaMA-\textit{X} decoder with sufficient information to generate clinically faithful reports, leading to a graceful degradation in performance rather than a catastrophic failure. 

\section{Experiments}

\begin{table*}[!t]
  \centering
  \caption{\textbf{Performance comparison} of our proposed DiA with state-of-the-art models on the IU X-Ray and MIMIC-CXR datasets, reporting NLG and CE metrics; Methods grouped as Image (Img), Knowledge-Guided (KG), \& Context-Aware (CA).}
  \vspace{-5pt}
  \label{tab:comparision_study}
  \begin{tabular}{llcccccccc}
    \toprule
    \multirow{2}{*}{\textbf{Type}} & \multirow{2}{*}{\textbf{Model}} & \multicolumn{4}{c}{\textbf{IU X-Ray}} & \multicolumn{4}{c}{\textbf{MIMIC-CXR}} \\ 
    \cmidrule(lr){3-6} \cmidrule(lr){7-10}
    & & \textbf{B@1} & \textbf{B@4} & \textbf{R-L} & \textbf{F$_1$} & \textbf{B@1} & \textbf{B@4} & \textbf{R-L} & \textbf{F$_1$} \\
    \midrule
    \multirow{2}{*}{\textbf{Img}} 
    & R2Gen \cite{chen2020generating} & 0.470 & 0.165 & 0.371 & - &  0.353 & 0.103 & 0.277 & -\\
    & CvT2Dis \cite{nicolson2023improving} & 0.473 & 0.175 & 0.376 & - & 0.392 & 0.127 & 0.285 & 0.384  \\
    \cmidrule(lr){1-10}
    \multirow{5}{*}{\textbf{KG}} & METransformer \cite{wang2023metransformer} & 0.483 & 0.172 & 0.380 & - & 0.386 & 0.124 & 0.291 & 0.311 \\
    &  Clinical BERT\cite{yan2022clinical} & 0.495 & 0.170 & 0.376 & - & 0.383 & 0.106 & 0.275 & 0.415  \\
    & M2KT \cite{yang2023radiology} & 0.497 & 0.174 & 0.399 & - & 0.386 & 0.111 & 0.274 & 0.352 \\
    &  MKSG \cite{yang2022knowledge} & 0.496 & 0.178 & 0.381 & - & 0.363 & 0.115 & 0.284 & 0.371  \\
    & XProNet \cite{wang2022cross} & 0.525 & 0.199 & 0.411 & - & 0.344 & 0.105 & 0.279 & - \\
    \cmidrule(lr){1-10}
    \multirow{3}{*}{\textbf{CA}}& PromptMRG \cite{jin2024promptmrg} & 0.401 & 0.098 & 0.281 & 0.211 & 0.398 & 0.112 & 0.268 & 0.476 \\
    & KiUT \cite{huang2023kiut} & 0.525 & 0.185 & 0.409 & - & 0.393 & 0.113 & 0.285 & 0.321 \\
    & EKAGen \cite{bu2024instance} & 0.526 & 0.203 & 0.404 & - & 0.411 & 0.119 & 0.217 & \textbf{0.499} \\
    & SEI \cite{liu2024structural} & - & - & - & - & 0.382 & \textbf{0.135} & 0.299 & 0.460 \\
    \cmidrule(lr){1-10}
    \multirow{1}{*}{\textbf{Ours}} & \textbf{DiA} & \textbf{0.616} & \textbf{0.266} & \textbf{0.516} & \textbf{0.298} & \textbf{0.415} & 0.134 & \textbf{0.369} & 0.497 \\
    \bottomrule
  \end{tabular}
  \vspace{-5pt}
\end{table*}

\begin{table*}[!t]
  \centering
  \caption{\textbf{Ablation Study}: Incremental effects of VL-MoE-VAE (\(\mathcal{L}_{\text{ELBO}}\)) and Disentangled Alignment (DA) (\(\mathcal{L}_{\text{orth}}+\mathcal{L}_{\text{align}}\)) across with-context (\cmark) and missing-context (\xmark) scenarios}
  \vspace{-5pt}
  \label{tab:ablation_study}
  \begin{tabular}{ccccccccccccc}
    \toprule
    \multirow{2}{*}{\textbf{Context}} &
    \multirow{2}{*}{\textbf{Baseline}} & \multirow{2}{*}{\textbf{VL-MoE-VAE}} & \multirow{2}{*}{\textbf{DA}} & \multicolumn{4}{c}{\textbf{IU X-Ray}} & \multicolumn{4}{c}{\textbf{MIMIC-CXR}} \\ 
    \cmidrule(lr){5-8} \cmidrule(lr){9-12}
    & & & & \textbf{B@1} & \textbf{B@4} & \textbf{R-L} & \textbf{F$_1$} & \textbf{B@1} & \textbf{B@4} & \textbf{R-L} & \textbf{F$_1$} \\
    \midrule
    \multirow{3}{*}{\cmark} & \cmark & \xmark & \xmark & 0.602 & 0.262 & 0.435 & 0.358 & 0.386 & 0.114 & 0.260 & 0.446 \\
    & \cmark & \cmark & \xmark & 0.655 & 0.319 & 0.548 & 0.381 & 0.423 & 0.140 & 0.343 & 0.551 \\
    & \cmark & \cmark & \cmark & \textbf{0.691} & \textbf{0.357} & \textbf{0.624} & \textbf{0.396} & \textbf{0.447} & \textbf{0.158} & \textbf{0.399} & \textbf{0.621} \\
    \midrule
    \multirow{3}{*}{\xmark} & \cmark & \xmark & \xmark & 0.276 & 0.079 & 0.185 & 0.166 & 0.295 & 0.049 & 0.176 & 0.219 \\
    & \cmark & \cmark & \xmark & 0.365 & 0.174 & 0.374 & 0.204 & 0.356 & 0.093 & 0.315 & 0.394 \\
    & \cmark & \cmark & \cmark & \textbf{0.387} & \textbf{0.198} & \textbf{0.421} & \textbf{0.213} & \textbf{0.371} & \textbf{0.104} & \textbf{0.350} & \textbf{0.438} \\
    \bottomrule
  \end{tabular}
  \vspace{-10pt}
\end{table*}

\subsection{Experimental Settings}
\subsubsection{Datasets and Preprocessing}
We evaluate DiA on two standard radiology report generation benchmarks: IU X-Ray~\cite{demner2016preparing} and MIMIC-CXR~\cite{johnson2019mimic}, both comprising paired chest X-ray images, free-text reports, and structured clinical metadata, enabling assessment under both complete and missing modality conditions. 

\textbf{IU X-Ray}, consists of 7,470 frontal-view X-ray images and 3,955 reports. We adopt a 70\%/10\%/20\% train/validation/test split and use a 1,000 word vocabulary. Approximately 2\% of the test samples in this dataset are missing clinical context, providing a controlled setting to test for modality resilience. \textbf{MIMIC-CXR} is a much larger dataset with 473,057 images and 206,563 reports across 64,588 patients. We use the official split from~\cite{chen2020generating}, comprising 270,790 training, 2,130 validation, and 3,858 test samples. Reports are tokenized, lower-cased, and filtered to remove non-alphabetic tokens; words appearing $<$ 4 are discarded, resulting in a vocabulary of 4,000 tokens. This dataset presents a more significant challenge for model robustness, as approximately 45\% of its test samples have missing clinical indications.

\subsubsection{Implementation and Training Details}
DiA was implemented in PyTorch and trained for 25 epochs on an NVIDIA A40 GPU using the AdamW optimizer~\cite{loshchilov2017decoupled} with a learning rate of 1e-4 and a weight decay of 1e-5. We used a batch size of 4 and set the maximum report length of 50 words. The model's compact architecture is defined by an embedding dimension $E$ of 1024, a latent dimension for $(Z_v, Z_l, Z_s)$ of 256, 6 Transformer encoder-decoder layers, 8 attention heads, and 2 key-value (KV) heads. A dropout rate of 0.1 was used to mitigate overfitting, while the loss term coefficients were set to $ \lambda_1, \lambda_2 = 0.3 $. These values were determined empirically from a search range of 0.1 to 0.5. To ensure consistent results, the Transformer's weight initialization was controlled by setting a random seed. We assess model performance using natural language generation (NLG) metrics including BLEU~\cite{papineni2002bleu} and ROUGE-L~\cite{lin2004rouge}, and a clinical efficacy (CE) metric such as F$_1$ score. Following~\cite{nicolson2023improving}, the F$_1$ score is calculated by converting the generated reports into 14 disease classification labels using the CheXbert labeler~\cite{smit2020chexbert}.

\begin{figure*}[!t]
    \centering
    \includegraphics[width=.95\textwidth]{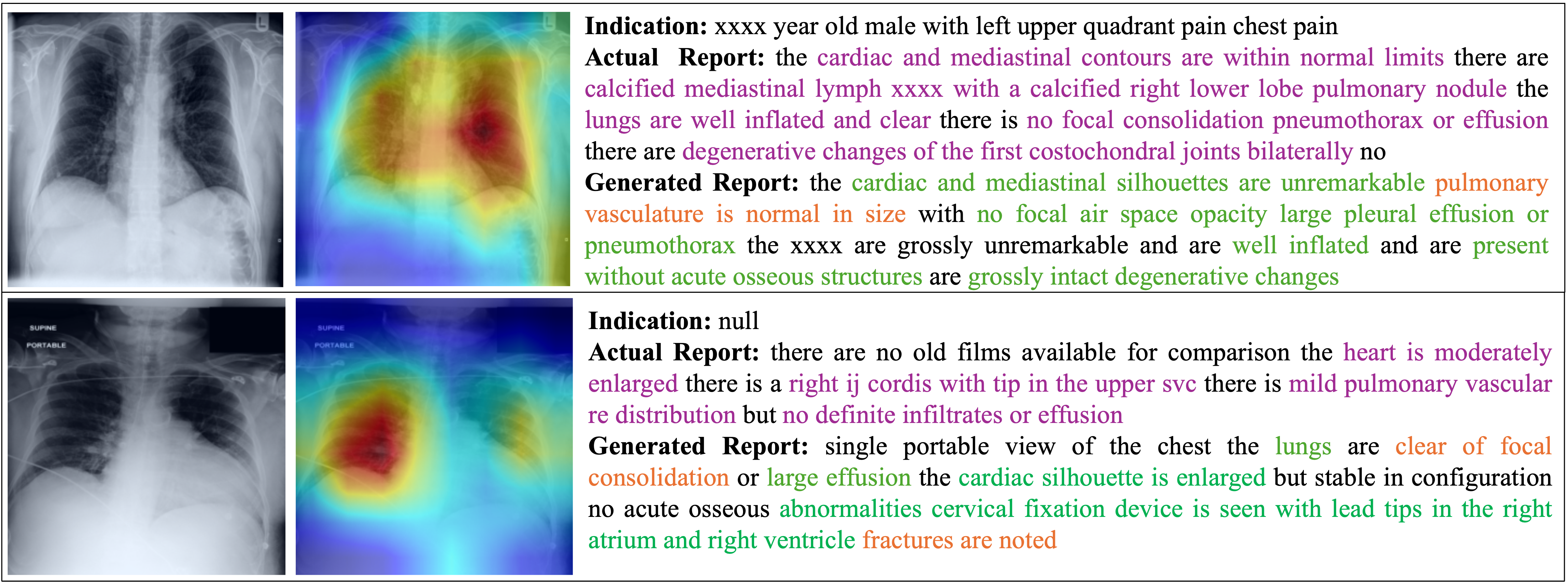}
    \vspace{-5pt}
    \caption{Comparison of actual and generated reports with chest X-rays and attention maps. Purple highlights key findings in the actual report, green indicates matched findings in the report, and amber marks mismatches / additional generated findings.}
    \label{fig:report_comparison}
    \vspace{-10pt}
\end{figure*}

\subsection{Evaluation}
\subsubsection{Comparison with State-of-the-Art Methods}
As shown in Table~\ref{tab:comparision_study}, DiA demonstrates superior performance compared to state-of-the-art (SOTA) methods on both IU X-Ray and MIMIC-CXR datasets. The evaluation spans Image-specific (Img), Knowledge-Guided (KG), and Context-Aware (CA) approaches, with DiA excelling in both natural language generation (NLG) and clinical efficacy (CE) metrics. On IU X-Ray, DiA achieves a BLEU@4 score of 0.266, surpassing the best KG model (XProNet) by 0.067, while an F$_1$ sccore of 0.298, outperming the best CA model (PromptMRG) by 0.087. On the more challenging MIMIC-CXR dataset, DiA's performance is highly competitive; while SEI shows a marginal lead in BLEU@4 (0.135 vs. 0.134), DiA’s higher ROUGE-L score indicates enhanced report coherence. Its F$_1$ score of 0.497 nearly matches the top performer, EKAGen (0.499). These results highlight DiA’s adept integration of vision-language contexts, surpassing advanced CA methods that struggle with longer reports. 

\subsubsection{Ablation Study: Impact of Core Components}
Table~\ref{tab:ablation_study} presents an ablation study quantifying the impact of DiA’s core components, the VL-MoE-VAE and the Disentangled Alignment (DA) constraint-under both complete (\cmark) and missing (\xmark) context scenarios. When clinical context is available, adding the VL-MoE-VAE to the baseline significantly boosts performance, improving the F$_1$ score on MIMIC-CXR by 0.105 and the BLEU@4 on IU X-ray by 0.057, which demonstrates the benefit of modeling a shared latent structure. Incorporating the DA constraint ($\mathcal{L}_{\text{orth}} + \mathcal{L}_{\text{align}}$) further enhances performance, with full DiA model achieving the highest scores across all metrics (e.g., MIMIC-CXR: F$_1$ 0.621, ROUGE-L 0.399). 

Under missing context, DiA shows remarkable resilience. While the baseline’s F$_1$ score on MIMIC-CXR drops by 0.227, with image only input, while DiA drops by only 0.183, outperforming the baseline by a margin of +0.219 in this challenging setting. The resilience is also evident on IU X-Ray, where DiA’s BLEU@4 remains more than 2$\times$ higher than baseline's (0.198 vs. 0.079). Comparing the start-to-end gains on MIMIC-CXR, the full DiA model improves over the baseline by 0.175 on the F$_1$ score with context and by 0.219 without context, demonstrating even greater relative benefit in the challenging incomplete-input scenario. 

These findings confirm that DiA’s latent structure effectively infers missing semantics, establishing DiA as a modality-resilient, high-performance report generator.

\begin{table}[t]
  \centering
  \caption{Comparison of encoder-decoder variants on MIMIC-CXR. RAD-DINO + CXR-BERT replaces DiA’s custom feature extractor and latent encoder; decoder across all variants is LLaMA-\textit{X} unless otherwise noted.}
  \vspace{-10pt}
  \label{tab:component_comparison}
  \begin{tabular}{lccccc}
    \toprule
    \textbf{Variant} & \textbf{Params} & \textbf{FLOPs} & \textbf{B@4} & \textbf{F$_1$} \\
    \midrule 
      RAD+CXR-BERT & 568.7 & 81.1 & 0.121 & 0.441 \\
    \midrule 
      Transformer & 591.2 & 80.6 & 0.126 & 0.479 \\
      GPT-2       & 746.9 & 86.4 & 0.116 & 0.419 \\
    \midrule
    \textbf{DiA LLaMA-\textit{X}} & \textbf{589.7} & \textbf{51.1} & \textbf{0.134} & \textbf{0.497} \\
    \bottomrule
  \end{tabular}
  \vspace{-10pt}
\end{table}

\subsubsection{Analysis of Architectural Choices and Efficiency}
Table~\ref{tab:component_comparison} summarizes a comparison of encoder and decoder variants on MIMIC-CXR to validate DiA's architectural design. For \textbf{encoder} variants, we compared DiA’s custom feature extraction pipeline against a pre-trained RAD-DINO + CXR-BERT setup.~\cite{perez2025exploring, boecking2022making} Despite using powerful pre-trained models, the RAD-DINO + CXR-BERT configuration achieved lower performance (BLEU@4 = 0.121, F$_1$ = 0.441) and incurred higher computational cost (81.1 GFLOPs). DiA's end-to-end learned encoder proved more effective and efficient (BLEU@4 = 0.134, F$_1$ = 0.497 at 51.1 GFLOPs).  For \textbf{decoder} variants, the LLaMA-\textit{X} architecture outperformed standard Transformer (BLEU@4 = 0.126, F$_1$ = 0.479 at 80.6 GFLOPs) and GPT-2 decoders (BLEU@4 = 0.116, F$_1$ = 0.419 at 86.4 GFLOPs) in both accuracy and efficiency. These results demonstrate that DiA's lightweight yet expressive components offer superior performance-to-cost trade-off.
DiA's efficiency is demonstrated by its training and inference times on an NVIDIA A40 GPU. Training on IU X-Ray takes 2.8 hours, with a  0.15-second inference time. For the larger MIMIC-CXR dataset, training takes 79.8 hours with a 0.18-second inference time. With 589.7M parameters and a computational cost of 51.14 GFLOPs, DiA maintains consistent computational efficiency.

\subsubsection{Qualitative Visual Inspection}
As shown in Figure~\ref{fig:report_comparison}, visual inspection of the model's attention maps reinforces its strengths. The heatmaps highlight that DiA focuses on key clinical regions in the chest X-rays, both with and without the presence of clinical context in the input. The strong alignment between the generated reports and the ground-truth reports underscores the effective synergy of all of DiA's components.

\section{Conclusion}
This research introduces DiA, a cutting-edge framework that advances radiology report generation by effectively integrating medical scans with real-time clinical indications. The core of DiA is its ability to disentangle and align modality-specific and shared latent representations, enabling the generation of coherent reports even with incomplete context. As a result, DiA outperforms state-of-the-art methods on the IU X-Ray and MIMIC-CXR datasets. This proven robustness in handling missing data underscores DiA’s potential to enhance diagnostic accuracy and support radiologists in real-world clinical scenarios. Overall, DiA significantly advances automating radiology reporting, promising to improve efficiency and reliability of medical imaging workflows.

\bibliography{aaai2026}

@article{chen2020generating,
  title={Generating radiology reports via memory-driven transformer},
  author={Chen, Zhihong and Song, Yan and Chang, Tsung-Hui and Wan, Xiang},
  journal={arXiv preprint arXiv:2010.16056},
  year={2020}
}

@article{yang2022knowledge,
  title={Knowledge matters: Chest radiology report generation with general and specific knowledge},
  author={Yang, Shuxin and Wu, Xian and Ge, Shen and Zhou, S Kevin and Xiao, Li},
  journal={Medical image analysis},
  volume={80},
  pages={102510},
  year={2022},
  publisher={Elsevier}
}

@inproceedings{yan2022clinical,
  title={Clinical-bert: Vision-language pre-training for radiograph diagnosis and reports generation},
  author={Yan, Bin and Pei, Mingtao},
  booktitle={Proceedings of the AAAI Conference on Artificial Intelligence},
  volume={36},
  pages={2982--2990},
  year={2022}
}

@article{yu2025adapter,
  title={Adapter-Enhanced Hierarchical Cross-Modal Pre-training for Lightweight Medical Report Generation},
  author={Yu, T. and Lu, W. and Yang, Y. and Han, W. and Huang, Q. and Yu, J. and Zhang, K.},
  journal={IEEE Journal of Biomedical and Health Informatics},
  year={2025}
}

@article{nicolson2023improving,
  title={Improving chest X-ray report generation by leveraging warm starting},
  author={Nicolson et. al, A.},
  journal={Artificial intelligence in medicine},
  volume={144},
  pages={102633},
  year={2023}
}

@inproceedings{wang2022cross,
  title={Cross-modal prototype driven network for radiology report generation},
  author={Wang, J. and Bhalerao, A. and He, Y.},
  booktitle={Computer Vision--ECCV 2022: 17th European Conference, Tel Aviv, Israel, October 23-27, 2022, Proceedings, Part XXXV},
  pages={563--579},
  year={2022},
  organization={Springer}
}

@inproceedings{wang2023metransformer,
  title={Metransformer: Radiology report generation by transformer with multiple learnable expert tokens},
  author={Wang, Z. and Liu, L. and Wang, L. and Zhou, L.},
  booktitle={Proceedings of the IEEE/CVF Conference on Computer Vision and Pattern Recognition},
  pages={11558--11567},
  year={2023}
}

@article{yang2023radiology,
  title={Radiology report generation with a learned knowledge base and multi-modal alignment},
  author={Yang, S. and Wu, X. and Ge, S. and Zheng, Z. and Zhou, S. K. and Xiao, L.},
  journal={Medical Image Analysis},
  volume={86},
  pages={102798},
  year={2023}
}

@inproceedings{huang2023kiut,
  title={Kiut: Knowledge-injected u-transformer for radiology report generation},
  author={Huang, Z. and Zhang, X. and Zhang, S.},
  booktitle={Proceedings of the IEEE/CVF Conference on Computer Vision and Pattern Recognition},
  pages={19809--19818},
  year={2023}
}

@inproceedings{li2023dynamic,
  title={Dynamic graph enhanced contrastive learning for chest x-ray report generation},
  author={Li, M. and Lin, B. and Chen, Z. and Lin, H. and Liang, X. and Chang, X.},
  booktitle={Proceedings of the IEEE/CVF Conference on Computer Vision and Pattern Recognition},
  pages={3334--3343},
  year={2023}
}

@inproceedings{bu2024instance,
  title={Instance-level expert knowledge and aggregate discriminative attention for radiology report generation},
  author={Bu, S. and Li, T. and Yang, Y. and Dai, Z.},
  booktitle={Proceedings of the IEEE/CVF Conference on Computer Vision and Pattern Recognition},
  pages={14194--14204},
  year={2024}
}

@inproceedings{jin2024promptmrg,
  title={Promptmrg: Diagnosis-driven prompts for medical report generation},
  author={Jin, H. and Che, H. and Lin, Y. and Chen, H.},
  booktitle={Proceedings of the AAAI Conference on Artificial Intelligence},
  volume={38},
  pages={2607--2615},
  year={2024}
}

@inproceedings{liu2024structural,
  title={Structural Entities Extraction and Patient Indications Incorporation for Chest X-Ray Report Generation},
  author={Liu, K. and Ma, Z. and Kang, X. and Zhong, Z. and Jiao, Z. and Baird, G. and Bai, H. and Miao, Q.},
  booktitle={International Conference on Medical Image Computing and Computer-Assisted Intervention},
  pages={433--443},
  year={2024},
  organization={Springer}
}

@inproceedings{yao2024drfuse,
  title={DrFuse: Learning Disentangled Representation for Clinical Multi-Modal Fusion with Missing Modality and Modal Inconsistency},
  author={Yao, Wenfang and Yin, Kejing and Cheung, William K. and Liu, Jia and Qin, Jing},
  booktitle={The Thirty-Eighth AAAI Conference on Artificial Intelligence (AAAI-24)},
  year={2024}
}

@article{huang2020fusion,
  title={Fusion of medical imaging and electronic health records using deep learning: a systematic review and implementation guidelines},
  author={Huang, S.-C. and Pareek, A. and Seyyedi, S. and Banerjee, I. and Lungren, M. P.},
  journal={NPJ Digital Medicine},
  volume={3},
  number={1},
  pages={136},
  year={2020}
}

@inproceedings{yoo2019deep,
  title={Deep learning of brain lesion patterns and user-defined clinical and MRI features for predicting conversion to multiple sclerosis from clinically isolated syndrome},
  author={Yoo, Y. and Tang, L. Y. and Li, D. K. and Metz, L. and Kolind, S. and Traboulsee, A. L. and Tam, R. C.},
  booktitle={Computer Methods in Biomechanics and Biomedical Engineering: Imaging \& Visualization},
  volume={7},
  pages={250--259},
  year={2019}
}

@inproceedings{ma2021smil,
  title={SMIL: Multimodal learning with severely missing modality},
  author={Ma, M. and Ren, J. and Zhao, L. and Tulyakov, S. and Wu, C. and Peng, X.},
  booktitle={Proceedings of the AAAI Conference on Artificial Intelligence},
  volume={35},
  pages={2302--2310},
  year={2021}
}

@inproceedings{sharma2019missing,
  title={Missing MRI pulse sequence synthesis using multi-modal generative adversarial network},
  author={Sharma, A. and Hamarneh, G.},
  booktitle ={IEEE Transactions on Medical Imaging},
  volume={39},
  pages={1170--1183},
  year={2019}
}

@inproceedings{braman2021deep,
  title={Deep Orthogonal Fusion: Multimodal Prognostic Biomarker Discovery Integrating Radiology, Pathology, Genomic, and Clinical Data},
  author={Braman, N. and Gordon, J. W. H. and Goossens, E. T. and Willis, C. and Stumpe, M. C. and Venkataraman, J.},
  booktitle={Medical Image Computing and Computer Assisted Intervention – MICCAI 2021},
  pages={667--677},
  year={2021}
}

@article{kline2022multimodal,
  title={Multimodal machine learning in precision health: A scoping review},
  author={Kline, A. and Wang, H. and Li, Y. and Dennis, S. and Hutch, M. and Xu, Z. and Wang, F. and Cheng, F. and Luo, Y.},
  journal={NPJ Digital Medicine},
  volume={5},
  number={1},
  pages={171},
  year={2022}
}

@article{cheerla2019deep,
  title={Deep learning with multimodal representation for pan-cancer prognosis prediction},
  author={Cheerla, A. and Gevaert, O.},
  journal={Bioinformatics},
  volume={35},
  number={14},
  pages={i446--i454},
  year={2019}
}

@article{huang2020multimodal,
  title={Multimodal fusion with deep neural networks for leveraging CT imaging and electronic health record: a case-study in pulmonary embolism detection},
  author={Huang, S.-C. and Pareek, A. and Zamanian, R. and Banerjee, I. and Lungren, M. P.},
  journal={Scientific Reports},
  volume={10},
  number={1},
  pages={22147},
  year={2020}
}

@article{venugopalan2021multimodal,
  title={Multimodal deep learning models for early detection of Alzheimer's disease stage},
  author={Venugopalan, J. and Tong, L. and Hassanzadeh, H. R. and Wang, M. D.},
  journal={Scientific reports},
  volume={11},
  number={1},
  pages={3254},
  year={2021}
}

@article{mohsen2022artificial,
  title={Artificial intelligence-based methods for fusion of electronic health records and imaging data},
  author={Mohsen, F. and Ali, H. and El Hajj, N. and Shah, Z.},
  journal={Scientific Reports},
  volume={12},
  number={1},
  pages={17981},
  year={2022}
}

@inproceedings{steyaert2023multimodal,
  title={Multimodal deep learning to predict prognosis in adult and pediatric brain tumors},
  author={Steyaert, S. and Qiu, Y. L. and Zheng, Y. and Mukherjee, P. and Vogel, H. and Gevaert, O.},
  booktitle={Communications Medicine},
  volume={3},
  pages={1--15},
  year={2023}
}

@inproceedings{zhang2022m3care,
  title={M3Care: Learning with missing modalities in multimodal healthcare data},
  author={Zhang, C. and Chu, X. and Ma, L. and Zhu, Y. and Wang, Y. and Wang, J. and Zhao, J.},
  booktitle={Proceedings of the 28th ACM SIGKDD Conference on Knowledge Discovery and Data Mining},
  pages={2418--2428},
  year={2022}
}

@article{li2023multi-modality,
  title={Multi-modality cardiac image computing: A survey},
  author={Li, L. and Ding, W. and Huang, L. and Zhuang, X. and Grau, V.},
  journal={Medical Image Analysis},
  volume={85},
  pages={102869},
  year={2023}
}

@inproceedings{sanchez2020learning,
  title={Learning Disentangled Representations via Mutual Information Estimation},
  author={Sanchez, E. H. and Serrurier, M. and Ortner, M.},
  booktitle={Computer Vision – ECCV 2020},
  pages={205--221},
  year={2020}
}

@inproceedings{chen2019robust,
  title={Robust multimodal brain tumor segmentation via feature disentanglement and gated fusion},
  author={Chen, C. and Dou, Q. and Jin, Y. and Chen, H. and Qin, J. and Heng, P.-A.},
  booktitle={Medical Image Computing and Computer Assisted Intervention – MICCAI 2019},
  pages={447--456},
  year={2019},
  organization={Springer}
}

@inproceedings{shen2019brain,
  title={Brain tumor segmentation on MRI with missing modalities},
  author={Shen, Y. and Gao, M.},
  booktitle={Information Processing in Medical Imaging: 26th International Conference, IPMI 2019},
  pages={417--428},
  year={2019},
  organization={Springer}
}

@inproceedings{cherukuri2024guided,
  author    = {Cherukuri, Teja Krishna and Shaik, Nagur Shareef and Ye, Dong Hye},
  title     = {Guided Context Gating: Learning to Leverage Salient Lesions in Retinal Fundus Images},
  booktitle = {Proceedings of the IEEE International Conference on Image Processing (ICIP)},
  year      = {2024},
  publisher = {IEEE},
  address   = {Abu Dhabi, United Arab Emirates},
  note      = {arXiv preprint arXiv:2406.13126}
}

@article{vaswani2017attention,
  title={Attention is all you need},
  author={Vaswani, Ashish and Shazeer, Noam and Parmar, Niki and Uszkoreit, Jakob and Jones, Llion and Gomez, Aidan N and Kaiser, {\L}ukasz and Polosukhin, Illia},
  journal={Advances in neural information processing systems},
  volume={30},
  year={2017}
}

@inproceedings{bousmalis2016domain,
  title={Domain separation networks},
  author={Bousmalis, Konstantinos and Trigeorgis, George and Silberman, Nathan and Krishnan, Dilip and Erhan, Dumitru},
  booktitle={Advances in Neural Information Processing Systems (NeurIPS)},
  year={2016},
  pages={343--351}
}

@article{touvron2023llama,
  title={Llama: Open and efficient foundation language models},
  author={Touvron, Hugo and Lavril, Thibaut and Izacard, Gautier and Martinet, Xavier and Lachaux, Marie-Anne and Lacroix, Timoth{\'e}e and Rozi{\`e}re, Baptiste and Goyal, Naman and Hambro, Eric and Azhar, Faisal and others},
  journal={arXiv preprint arXiv:2302.13971},
  year={2023}
}

@article{brown2020language,
  title={Language models are few-shot learners},
  author={Brown, Tom B},
  journal={arXiv preprint arXiv:2005.14165},
  year={2020}
}

@article{rusak2024infonce,
  title={InfoNCE: Identifying the Gap Between Theory and Practice},
  author={Rusak, Evgenia and Reizinger, Patrik and Juhos, Attila and Bringmann, Oliver and Zimmermann, Roland S and Brendel, Wieland},
  journal={arXiv preprint arXiv:2407.00143},
  year={2024}
}

@inproceedings{mao2023multimodal,
  title={Multimodal variational auto-encoder based audio-visual segmentation},
  author={Mao, Yuxin and Zhang, Jing and Xiang, Mochu and Zhong, Yiran and Dai, Yuchao},
  booktitle={Proceedings of the IEEE/CVF International Conference on Computer Vision},
  pages={954--965},
  year={2023}
}

@article{menendez1997jensen,
  title={The jensen-shannon divergence},
  author={Men{\'e}ndez, Mar{\'\i}a Luisa and Pardo, Julio Angel and Pardo, Leandro and Pardo, Mar{\'\i}a del C},
  journal={Journal of the Franklin Institute},
  volume={334},
  number={2},
  pages={307--318},
  year={1997},
  publisher={Elsevier}
}

@article{demner2016preparing,
  title={Preparing a collection of radiology examinations for distribution and retrieval},
  author={Demner-Fushman, Dina and Kohli, Marc D and Rosenman, Marc B and Shooshan, Sonya E and Rodriguez, Laritza and Antani, Sameer and Thoma, George R and McDonald, Clement J},
  journal={Journal of the American Medical Informatics Association},
  volume={23},
  number={2},
  pages={304--310},
  year={2016},
  publisher={Oxford University Press}
}

@article{johnson2019mimic,
  title={MIMIC-CXR-JPG, a large publicly available database of labeled chest radiographs},
  author={Johnson, Alistair EW and Pollard, Tom J and Greenbaum, Nathaniel R and Lungren, Matthew P and Deng, Chih-ying and Peng, Yifan and Lu, Zhiyong and Mark, Roger G and Berkowitz, Seth J and Horng, Steven},
  journal={arXiv preprint arXiv:1901.07042},
  year={2019}
}

@inproceedings{tan2019efficientnet,
  title={Efficientnet: Rethinking model scaling for convolutional neural networks},
  author={Tan, Mingxing and Le, Quoc},
  booktitle={International conference on machine learning},
  pages={6105--6114},
  year={2019},
  organization={PMLR}
}

@article{simonyan2014very,
  title={Very deep convolutional networks for large-scale image recognition},
  author={Simonyan, Karen and Zisserman, Andrew},
  journal={arXiv preprint arXiv:1409.1556},
  year={2014}
}

@inproceedings{liu2019transformer,
  title={A transformer-based variational autoencoder for sentence generation},
  author={Liu, Danyang and Liu, Gongshen},
  booktitle={2019 International Joint Conference on Neural Networks (IJCNN)},
  pages={1--7},
  year={2019},
  organization={IEEE}
}

@article{loshchilov2017decoupled,
  title={Decoupled weight decay regularization},
  author={Loshchilov, Ilya and Hutter, Frank},
  journal={arXiv preprint arXiv:1711.05101},
  year={2017}
}

@inproceedings{papineni2002bleu,
  title={Bleu: a method for automatic evaluation of machine translation},
  author={Papineni, Kishore and Roukos, Salim and Ward, Todd and Zhu, Wei-Jing},
  booktitle={Proceedings of the 40th annual meeting of the Association for Computational Linguistics},
  pages={311--318},
  year={2002}
}

@inproceedings{lin2004rouge,
  title={Rouge: A package for automatic evaluation of summaries},
  author={Lin, Chin-Yew},
  booktitle={Text summarization branches out},
  pages={74--81},
  year={2004}
}

@inproceedings{smit2020chexbert,
    title={Combining Automatic Labelers and Expert Annotations for Accurate Radiology Report Labeling Using BERT},
    author={Smit, Akshay  and Jain, Saahil  and Rajpurkar, Pranav  and Pareek, Anuj  and Ng, Andrew  and Lungren, Matthew},
    booktitle={Proceedings of the 2020 Conference on Empirical Methods in Natural Language Processing (EMNLP)},
    month = nov,
    year={2020},
    publisher={Association for Computational Linguistics},
    pages={1500--1519},
}

@article{shi2019variational,
  title={Variational mixture-of-experts autoencoders for multi-modal deep generative models},
  author={Shi, Yuge and Paige, Brooks and Torr, Philip and others},
  journal={Advances in neural information processing systems},
  volume={32},
  year={2019}
}

@article{poole2019variational,
  title={Variational Inference with Mutual Information Constraints},
  author={Poole, Ben and van den Oord, Aaron and Hjelm, R Devon and Maal{\o}e, Lars and Dhariwal, Prafulla and Kingma, Diederik P and Alemi, Alexander A},
  journal={arXiv preprint arXiv:1907.00030},
  year={2019}
}

@inproceedings{hayat2022medfuse,
  title={MedFuse: Multi-modal fusion with clinical time-series data and chest X-ray images},
  author={Hayat, N. and Geras, K. J. and Shamout, F. E.},
  booktitle={Proceedings of the 7th Machine Learning for Healthcare Conference},
  volume={182},
  pages={479--503},
  year={2022},
  organization={PMLR}
}

@inproceedings{drfuse2024,
  title={{DrFuse}: Learning Disentangled Representation for Clinical Multi-Modal Fusion with Missing Modality and Modal Inconsistency},
  author={Yao, Wenfang and Yin, Kejing and Cheung, William K. and Liu, Jia and Qin, Jing},
  booktitle={The Thirty-Eighth AAAI Conference on Artificial Intelligence (AAAI-24)},
  year={2024}
}

@article{drim2024,
  title={DRIM: Learning Disentangled Representations from Incomplete Multimodal Healthcare Data},
  author={Robinet, Lucas and Berjaoui, Ahmad and Kheil, Ziad and Cohen-Jonathan Moyal, Elizabeth},
  journal={arXiv preprint arXiv:2409.17055},
  year={2024}
}

@article{imdr2025,
  title={Incomplete Modality Disentangled Representation for Ophthalmic Disease Grading and Diagnosis},
  author={Liu, Chengzhi and Huang, Zile and Chen, Zhe and Tang, Feilong and Tian, Yu and Xu, Zhongxing and Luo, Zihong and Zheng, Yalin and Meng, Yanda},
  journal={arXiv preprint arXiv:2502.11724},
  year={2025}
}

@article{wu2018multimodal,
  title={Multimodal generative models for scalable weakly-supervised learning},
  author={Wu, Mike and Goodman, Noah},
  journal={Advances in neural information processing systems},
  volume={31},
  year={2018}
}

@article{minka2005divergence,
  title={Divergence measures and message passing},
  author={Minka, Tom and others},
  year={2005},
  publisher={Technical report, Microsoft Research},
  journal={Technical Report MSR-TR-2005-173}
}

@article{sutter2020multimodal,
  title={Multimodal generative learning utilizing jensen-shannon-divergence},
  author={Sutter, Thomas and Daunhawer, Imant and Vogt, Julia},
  journal={Advances in neural information processing systems},
  volume={33},
  pages={6100--6110},
  year={2020}
}

@incollection{hyvarinen2001independent,
  title={Independent component analysis},
  author={Hyv{\"a}rinen, Aapo and Hurri, Jarmo and Hoyer, Patrik O},
  booktitle={Natural Image Statistics: A Probabilistic Approach to Early Computational Vision},
  pages={151--175},
  year={2001},
  publisher={Springer}
}

@article{perez2025exploring,
  title={Exploring scalable medical image encoders beyond text supervision},
  author={Perez-Garcia, Fernando and Sharma, Harshita and Bond-Taylor, Sam and Bouzid, Kenza and Salvatelli, Valentina and Ilse, Maximilian and Bannur, Shruthi and Castro, Daniel C and Schwaighofer, Anton and Lungren, Matthew P and others},
  journal={Nature Machine Intelligence},
  volume={7},
  number={1},
  pages={119--130},
  year={2025},
  publisher={Nature Publishing Group UK London}
}

@inproceedings{boecking2022making,
  title={Making the most of text semantics to improve biomedical vision--language processing},
  author={Boecking, Benedikt and Usuyama, Naoto and Bannur, Shruthi and Castro, Daniel C and Schwaighofer, Anton and Hyland, Stephanie and Wetscherek, Maria and Naumann, Tristan and Nori, Aditya and Alvarez-Valle, Javier and others},
  booktitle={European conference on computer vision},
  pages={1--21},
  year={2022},
  organization={Springer}
}

\clearpage

\appendix
\section{Supplementary Material}

\subsection{Derivation of the Evidence Lower Bound (ELBO)}

In this section, we provide a detailed derivation of the Evidence Lower Bound (ELBO) objective optimized by our DiA-gnostic VLVAE. Our model leverages a tri-factor latent space consisting of modality-specific variables $Z_v$, $Z_l$ and a shared latent variable $Z_s$, constrained via disentanglement and alignment regularizers. Importantly, the posterior over $Z_s$ is formulated as a Mixture-of-Experts (MoE), which enables robust inference under missing modalities. The generative model assumes the following factorized structure:
\begin{align*}
    p_\theta(V, L, Z_v, Z_l, Z_s) =& p_\theta(V \mid Z_v)\, p_\theta(L \mid Z_l)\, \\ \notag
                                    &p(Z_v)\, p(Z_l)\, p(Z_s) \notag
\end{align*}
Here, $Z_v$ and $Z_l$ are modality-specific latent variables for vision $V$ and language $L$, respectively. The shared latent $Z_s \sim \mathcal{N}(0, I)$ encodes cross-modal semantics, and while it is not used directly in the decoders, it is regulated via auxiliary constraints. This simplifies decoding while allowing $Z_s$ to influence training through alignment objectives and cross-modal supervision.

We seek to maximize the marginal log-likelihood $\log p_\theta(V, L)$, which is lower bounded via:
\begin{align*}
    \log& p_\theta(V, L) = \mathcal{L}_{\text{ELBO}} + \\ \notag
        & D_{\text{KL}}(q_\phi(Z_v, Z_l, Z_s \mid V, L) \| p_\theta(Z_v, Z_l, Z_s \mid V, L))
\end{align*}
This implies:
\begin{align*}
    \mathcal{L}_{\text{ELBO}} = \mathbb{E}_{q_\phi} \left[ \log \frac{p_\theta(V, L, Z_v, Z_l, Z_s)}{q_\phi(Z_v, Z_l, Z_s \mid V, L)} \right]
\end{align*}
The approximate posterior factorizes as: 
\begin{align*}
q_\phi(Z_v, Z_l, Z_s \mid V, L) = q_{\phi_v}(Z_v \mid V)\, q_{\phi_l}(Z_l \mid L)\, q_{\phi_s}(Z_s \mid V, L)
\end{align*}
where $q_{\phi_s}(Z_s \mid V, L)$ is implemented as a mixture of modality-specific experts. Substituting the factorized distributions, we expand $\mathcal{L}_{\text{ELBO}}$ as:
\begin{align}
    \mathcal{L}_{\text{ELBO}} &=\notag\mathbb{E}_{q_{\phi_v}, q_{\phi_l}, q_{\phi_s}} \big[ \log p_\theta(V \mid Z_v) + \log p_\theta(L \mid Z_l) \big] \\ \notag
    &+ \mathbb{E}_{q_{\phi_v}} [\log p(Z_v) - \log q_{\phi_v}(Z_v \mid V)] \\ \notag
    &+ \mathbb{E}_{q_{\phi_l}} [\log p(Z_l) - \log q_{\phi_l}(Z_l \mid L)] \\ \notag
    &+ \mathbb{E}_{q_{\phi_s}} [\log p(Z_s) - \log q_{\phi_s}(Z_s \mid V, L)]     
\end{align}
Grouping terms yields:
\begin{align*}
    \mathcal{L}_{\text{ELBO}} &= \mathbb{E}_{q_{\phi_s}(Z_s \mid V, L)} \Big[\mathbb{E}_{q_{\phi_v}(Z_v \mid V)} \log p_{\theta_v}(V \mid Z_v)\Big] \\  
    &+ \mathbb{E}_{q_{\phi_s}(Z_s \mid V, L)}\Big[\mathbb{E}_{q_{\phi_l}(Z_l \mid L)} \log p_{\theta_l}(L \mid Z_l)
    \Big] \notag \\
    &- D_{\text{KL}}(q_{\phi_v}(Z_v \mid V) \| p(Z_v))  \notag \\
    &- D_{\text{KL}}(q_{\phi_l}(Z_l \mid L) \| p(Z_l)) \notag \\
    &- D_{\text{KL}}(q_{\phi_s}(Z_s \mid V, L) \| p(Z_s))
\end{align*}
In our formulation, the shared posterior $q_{\phi_s}(Z_s \mid V, L)$ is a mixture of unimodal experts:
\begin{align*}
    q_{\phi_s}(Z_s \mid V, L) = \pi_v q_{\phi_s}(Z_s \mid V) + \pi_l q_{\phi_s}(Z_s \mid L)
\end{align*}
where $\pi_v, \pi_l$ are data-dependent mixture weights. This mixture distribution may not be absolutely continuous with respect to $p(Z_s)$, which causes instability when computing the KL divergence. Following prior work in multimodal VAEs, we replace the KL term with the Jensen-Shannon Divergence:
\begin{align*}
    \text{JSD}(q_{\phi_s}(Z_s \mid V, L)\, \|\, p(Z_s))
\end{align*}

This leads to the final training objective as in eq. (4):
\begin{align*}
    \mathcal{L}_{\text{ELBO}} &= \mathbb{E}_{q_{\phi_s}(Z_s \mid V, L)} \Big[\mathbb{E}_{q_{\phi_v}(Z_v \mid V)} \log p_{\theta_v}(V \mid Z_v)\Big] \notag \\
    &+ \mathbb{E}_{q_{\phi_s}(Z_s \mid V, L)}\Big[\mathbb{E}_{q_{\phi_l}(Z_l \mid L)} \log p_{\theta_l}(L \mid Z_l)\Big] \notag \notag \\
    &- D_{\text{KL}}(q_{\phi_v}(Z_v \mid V) \| p(Z_v)) \notag \\
    &- D_{\text{KL}}(q_{\phi_l}(Z_l \mid L) \| p(Z_l)) \notag \\
    &- \text{JSD}(q_{\phi_s}(Z_s \mid V, L) \, \|\, p(Z_s))
    \label{eq:elbo_final}
\end{align*}

This ELBO objective is used during training to learn a disentangled, semantically aligned latent representation across modalities. Although $Z_s$ is not directly used in the reconstruction paths, it plays a vital role in enforcing global semantic consistency and enables robust inference under missing modality conditions.

\subsection{Disentangled Alignment Constraints}

To encourage a semantically structured latent space, DiA-gnostic VLVAE introduces two complementary regularization losses: an orthogonality constraint to enforce statistical disentanglement, and a contrastive alignment loss to ensure cross-modal consistency.

\subsubsection*{Proposition 1 (Disentanglement via Orthogonality)}

\textit{Let $(\tilde{Z}_s, \tilde{Z}_v, \tilde{Z}_l)$ be whitened latent vectors with zero mean and unit variance. If the decoder is locally linear in latent space, then minimizing the following orthogonality loss:}
\[
\mathcal{L}_{\text{orth}} = \| \tilde{Z}_s^\top \tilde{Z}_v \|_F^2 + \| \tilde{Z}_s^\top \tilde{Z}_l \|_F^2 + \| \tilde{Z}_v^\top \tilde{Z}_l \|_F^2
\]
\textit{encourages all latent factors to be mutually uncorrelated. Under the assumptions of Independent Component Analysis (ICA), such uncorrelatedness implies statistical independence.}

\paragraph{Proof Sketch.} The Frobenius norm $\| X^\top Y \|_F^2$ measures the sum of squared pairwise covariances between components of $X$ and $Y$. For whitened vectors (zero mean and unit variance), these terms reduce to:
\[
\| \tilde{Z}_s^\top \tilde{Z}_v \|_F^2 \propto \sum_{i,j} \text{Cov}^2(\tilde{Z}_{s,i}, \tilde{Z}_{v,j})
\]
and analogously for the other pairs. Minimizing $\mathcal{L}_{\text{orth}}$ to zero enforces that all pairwise covariances vanish, i.e., that $Z_s$, $Z_v$, and $Z_l$ are mutually uncorrelated. Under ICA assumptions, this guarantees statistical independence of the latent components.

\begin{figure*}[t]
    \centering
    \includegraphics[width=\textwidth]{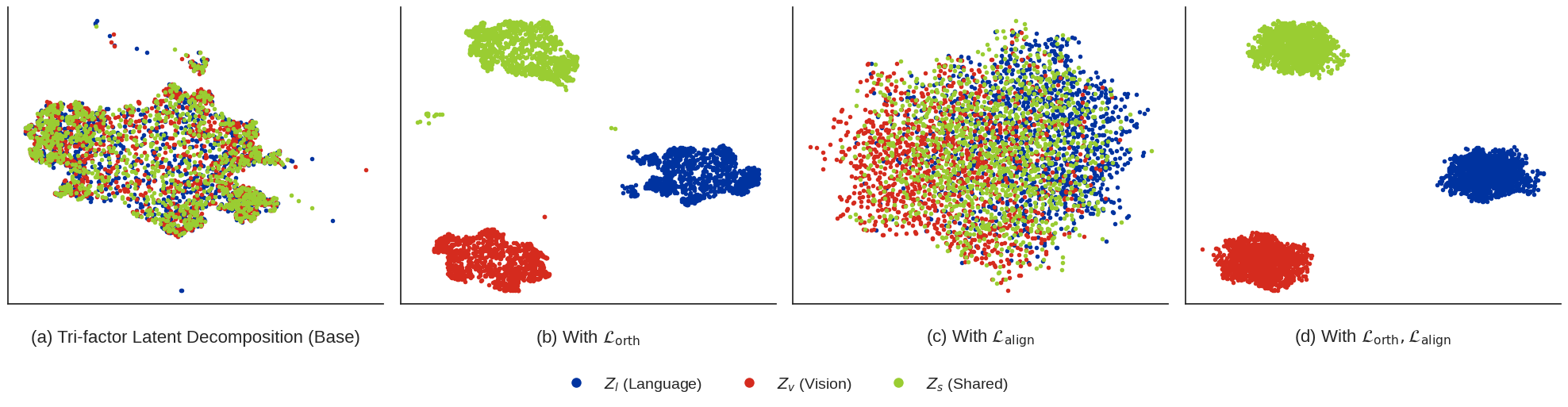}    
    \caption{t-SNE projections of latent variables for IU X-Ray. Each subfigure shows distributions of language-specific (\(Z_l\), blue), vision-specific (\(Z_v\), red), and shared (\(Z_s\), green) representations under four settings: (a) Base VLVAE, (b) with \(\mathcal{L}_{\text{orth}}\), (c) with \(\mathcal{L}_{\text{align}}\), and (d) with both constraints.}
    \label{fig:tsne-iux}
    \vspace{-10pt}
\end{figure*}

\begin{figure*}[t]
    \centering
    \includegraphics[width=\textwidth]{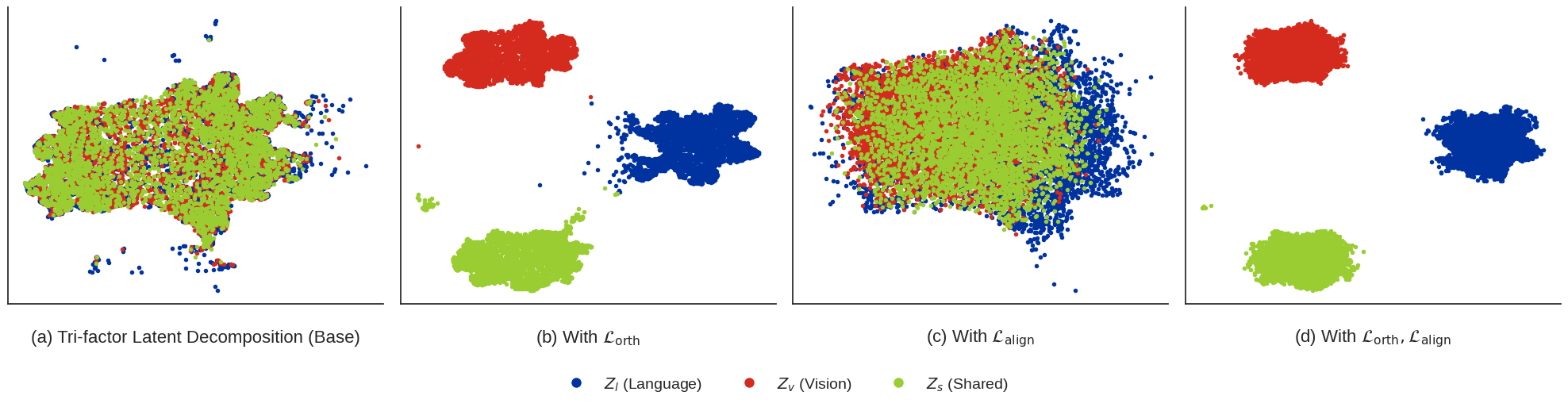}
    \caption{t-SNE projections of latent variables for MIMIC-CXR under the same settings as in Fig.~\ref{fig:tsne-iux}. The plots illustrate how the latent space evolves across training objectives and datasets.}
    \label{fig:tsne-mimic}
    \vspace{-10pt}
\end{figure*}

\subsubsection*{Proposition 2 (Alignment via Contrastive Loss)}

\textit{Let $Z_s$ be the shared latent representation and $Z_v$, $Z_l$ the modality-specific latents. Then minimizing the contrastive alignment loss:}
\[
\mathcal{L}_{\text{align}} = \mathcal{L}_{\text{align}}^{(v)} + \mathcal{L}_{\text{align}}^{(l)}
\]
\textit{where each term is defined using the InfoNCE objective, e.g.,}
\[
\mathcal{L}_{\text{align}}^{(v)} = -\mathbb{E} \left[ \log \frac{\exp(\text{sim}(Z_s, Z_v)/\tau)}{\sum_{Z_v'} \exp(\text{sim}(Z_s, Z_v')/\tau)} \right],
\]
\textit{maximizes a variational lower bound on the mutual information $I(Z_s; Z_v)$ and $I(Z_s; Z_l)$, respectively.}

\paragraph{Proof Sketch.} The InfoNCE objective with $K-1$ negative samples satisfies:
\[
I(X; Y) \ge \log K - \mathcal{L}_{\text{NCE}}
\]
Therefore, minimizing $\mathcal{L}_{\text{align}}^{(v)}$ increases a lower bound on $I(Z_s; Z_v)$, encouraging the shared latent to retain relevant modality-specific semantics. The same argument applies to $\mathcal{L}_{\text{align}}^{(l)}$.

\paragraph{Remark.} Together, $\mathcal{L}_{\text{orth}}$ and $\mathcal{L}_{\text{align}}$ enable the model to learn a disentangled yet semantically grounded latent representation that generalizes across modality configurations.

\subsection{Latent Structure Visualization}

\paragraph{Latent Disentanglement and Alignment Analysis.}
We visualize the latent distributions of modality-specific (\(Z_v\), \(Z_l\)) and shared (\(Z_s\)) representations using t-SNE on IUXRay (Figure \ref{fig:tsne-iux}) and MIMIC-CXR (Figure \ref{fig:tsne-mimic}) to assess the impact of the disentangled alignment constraints \(\mathcal{L}_{\text{orth}}\) and \(\mathcal{L}_{\text{align}}\). Without any constraints, the base model exhibits heavy entanglement across all latents, indicating poor separation of modality-specific and shared semantics. Introducing only \(\mathcal{L}_{\text{orth}}\) yields clear separation among \(Z_v\), \(Z_l\), and \(Z_s\), effectively disentangling modality-specific features. However, the shared latent remains misaligned, lacking semantic coherence. Conversely, \(\mathcal{L}_{\text{align}}\) alone collapses all representations into a semantically aligned cluster but compromises disentanglement by blurring modality-specific distinctions.

When both constraints are applied jointly, the resulting latent structure achieves optimal balance: \(Z_v\) and \(Z_l\) form well-separated clusters, while \(Z_s\) aligns closely with both, indicating successful capture of shared semantics without sacrificing modality identity. This structured organization confirms that the proposed disentangled alignment not only enforces statistical independence via orthogonality but also encourages semantic consistency through contrastive alignment. The consistency of this effect across both datasets highlights DiA’s generalization capability and supports its core contribution: learning modality-resilient, interpretable latent spaces for robust cross-modal report generation.

\begin{table*}[ht]
\centering
\caption{Architectural Specifications of DiA Components.}
\vspace{-5pt}
\label{tab:arch-specs}
\begin{tabular}{|l|l|p{6cm}|}
\hline
\textbf{Component} & \textbf{Base Model / Type} & \textbf{Details} \\
\hline
Vision Feature Extractor & EfficientNetB0 & Pre-trained on ImageNet; appended with a Global Context Attention (GCA) module. Output dim: 1024. \\ \hline
Language Feature Extractor & Transformer Encoder & 6 layers, 8 heads, FF dim 2048, GELU, dropout 0.1. \\ \hline
Modality Abstractor & Bidirectional Cross-Attention & 2 layers, 8 heads \\
\hline
\multicolumn{3}{|c|}{\textbf{VL-MoE-VAE Encoders}} \\
\hline
Vision-Specific ($q_{\phi_v}$) & VGG16 + MLP & Pre-trained on ImageNet; final conv features fed to 2-layer MLP for $\mu_v, \sigma_v$. \\ \hline
Language-Specific ($q_{\phi_l}$) & Transformer Encoder & 4 layers, 8 heads, FF dim 1024; outputs $\mu_l, \sigma_l$. \\ \hline
Shared Encoder ($q_{\phi_s}$) & MLP (MoE) & Two 2-layer expert MLPs (vision/language), hidden size 512; outputs $\mu_s, \sigma_s$. \\
\hline
\multicolumn{3}{|c|}{\textbf{VL-MoE-VAE Decoders}} \\
\hline
Vision Decoder ($p_{\theta_v}$) & Transposed CNN & 5-layer transposed conv network.  \\ \hline 
Language Decoder ($p_{\theta_l}$) & Transformer Decoder & 4 layers, 8 heads. \\ \hline 
LLaMA-X Decoder & Transformer Decoder & 6 layers, 8 heads, 2 KV heads, SwiGLU, RoPE positional encoding. \\
\hline
\end{tabular}
\end{table*}

\subsection{Marginal ELBO under Missing Language Context}

A critical feature of the DiA framework is its ability to handle incomplete data, a common scenario in clinical settings where textual context $L$ may be unavailable during inference. The Mixture-of-Experts (MoE) design provides a principled way to manage this by allowing the model to fall back on unimodal inference from the available vision data. This section details the derivation of the marginal Evidence Lower Bound (ELBO) that justifies this process, demonstrating that the framework remains theoretically sound even with partial inputs.

When the language modality $L$ is missing (e.g., passed as a NULL token), the MoE router learns to down-weight the corresponding expert, effectively conditioning the shared posterior only on the vision input. Our goal is to show that the learning objective remains a valid lower bound on the marginal log-likelihood of the observed vision data, $\log p_\theta(V)$. The derivation begins with the definition of the marginal log-likelihood and its relationship to the ELBO:
\begin{align*}
    \log p_\theta(V) &= \log \iint p_\theta(V, Z_v, Z_s)\, dZ_v\, dZ_s \\
    &\ge \mathbb{E}_{q_\phi(Z_v, Z_s \mid V)} \left[ \log \frac{p_\theta(V, Z_v, Z_s)}{q_\phi(Z_v, Z_s \mid V)} \right]
\end{align*}

Assuming the posterior factorizes as $q_\phi(Z_v, Z_s \mid V)=q_{\phi_v}(Z_v \mid V)\, q_{\phi_s}(Z_s \mid V)$ and using the generative factorization $p_\theta(V, Z_v, Z_s) = p_\theta(V \mid Z_v)\, p(Z_v)\, p(Z_s)$, we expand the objective. This expansion yields the final marginal ELBO for the vision-only case:
\begin{align*}
    \mathcal{L}_{\text{ELBO}}^{(V)} = &\; \mathbb{E}_{q_{\phi_v}(Z_v \mid V)} \big[ \log p_\theta(V \mid Z_v) \big] \\
    &- D_{\text{KL}}\big( q_{\phi_v}(Z_v \mid V) \,\|\, p(Z_v) \big) \\
    &- \text{JSD}\big( q_{\phi_s}(Z_s \mid V) \,\|\, p(Z_s) \big)
\end{align*}

In this case, the posterior over $Z_s$ reduces to the vision-specific expert, i.e., $q_{\phi_s}(Z_s \mid V)$, since $\pi_l = 0$ and $\pi_v = 1$ in the MoE formulation:
\begin{align*}
    q_{\phi_s}(Z_s \mid V, \text{NULL}) = q_{\phi_s}(Z_s \mid V)
\end{align*}

Therefore, even in the absence of language modality, the DiA framework yields a valid and optimized ELBO for unimodal input. This marginal ELBO retains semantic structure through $Z_s$ and enables modality-resilient report generation without requiring explicit imputation or retraining.

\begin{table*}[ht]
\centering
\caption{Training hyperparameters and computational environment.}
\vspace{-5pt}
\label{tab:training-specs}
\begin{tabular}{|l|l|}
\hline
\textbf{Parameter} & \textbf{Value / Description} \\
\hline
Optimizer & AdamW ($\beta_1=0.9$, $\beta_2=0.999$, $\epsilon=10^{-8}$) \\
Learning Rate & $1\times10^{-4}$ with linear warmup \\
Weight Decay & $1\times10^{-5}$ (excluding bias and LayerNorm) \\
Batch Size & 4 \\
Epochs & 25 \\
Latent Dim ($Z_v, Z_l, Z_s$) & 256 \\
Embedding Dim ($E$) & 1024 \\
$\lambda_1$ (Orthogonality) & 0.3 (search: \{0.1, 0.3, 0.5\}) \\
$\lambda_2$ (Alignment) & 0.3 (search: \{0.1, 0.3, 0.5\}) \\
Temperature ($\tau$) & 0.07 (InfoNCE loss) \\
GPU & 1x NVIDIA A40 (48GB) \\
Software & PyTorch 2.1, CUDA 12.1 \\
\hline
\end{tabular}
\end{table*}

\begin{table*}[ht]
\centering
\caption{Statistics for IU X-Ray and MIMIC-CXR datasets.}
\vspace{-5pt}
\label{tab:dataset-stats}
\begin{tabular}{|l|c|c|c|c|c|c|}
\hline
\textbf{Dataset} & Train & Val & Test & Vocab Size & Avg. Len. & \% Missing (Test) \\
\hline
IU X-Ray & 5,229 & 747 & 1,501 & $\sim$1,000 & 33 & $\sim$2\% \\
MIMIC-CXR & 270,790 & 2,130 & 3,858 & $\sim$4,000 & 58 & $\sim$45\% \\
\hline
\end{tabular}
\end{table*}

\subsection{Implementation and Architectural Details}
All models were implemented in PyTorch, and the source code has been provided for full reproducibility. The following sections detail the model architectures, training hyperparameters, and dataset statistics, with specific configurations summarized in the corresponding tables.

\subsubsection{Model Architectures}
The detailed configurations of the core DiA components are specified in Table~\ref{tab:arch-specs}. The framework uses pre-trained backbones for initial feature extraction, including an EfficientNetB0 for the vision extractor and a 6-layer Transformer for the language extractor. For the probabilistic VL-MoE-VAE module, the modality-specific encoders consist of a VGG16+MLP for vision and a 4-layer Transformer for language. The final report generation uses a 6-layer LLaMA-X decoder, which is optimized with features like SwiGLU activation and RoPE positional encodings.

\subsubsection{Training Hyperparameters}
The training hyperparameters and computational environment are summarized in Table~\ref{tab:training-specs}.
All models were trained for 25 epochs with a batch size of 4 using the AdamW optimizer. 
We used a learning rate of $1\times10^{-4}$ with a linear warmup for the first 10\% of training steps. The latent and embedding dimensions are 256 and 1024, respectively. The loss term coefficients $\lambda_1$ (Orthogonality)  and $\lambda_2$ (Alignment), were both set to 0.3 after a search over the set {0.1, 0.3, 0.5}. The experiments were conducted on a single NVIDIA A40 GPU using PyTorch 2.1.

\subsubsection{Dataset Statistics}
Table~\ref{tab:dataset-stats} provides the key statistics for the IU X-Ray and MIMIC-CXR datasets used in our experiments. The statistics include the train, validation, and test splits, as well as the vocabulary size and average report length for each dataset. Notably, the table highlights the difference in data scarcity between the two benchmarks, with the MIMIC-CXR test set having a significantly higher rate of missing clinical context ($\sim$45\%) compared to IU X-Ray ($\sim$2\%).

\end{document}